%% file: iclr2024_conference.tex
\documentclass{article} 
\usepackage{iclr2024_conference,times}
\usepackage{hyperref}
\usepackage{url}
\usepackage{booktabs}
\usepackage{tabularx}
\usepackage{makecell}
\usepackage{array}
\usepackage{caption}
\usepackage{graphicx}
\usepackage{enumitem}
\usepackage{wrapfig}
\usepackage{lipsum}
\usepackage{amssymb}
\usepackage{pifont}
\usepackage{xcolor}
\usepackage{multirow}
\usepackage{float}
\definecolor{OliveGreen}{rgb}{0.33, 0.42, 0.18}
\definecolor{positive}{HTML}{1E90FF} 
\definecolor{negative}{HTML}{FF6347} 

\makeatletter
\def\blfootnote{\gdef\@thefnmark{}\@footnotetext}
\makeatother

\input{math_commands.tex}

\title{Improving LLM Reasoning through Scaling Inference Computation with Collaborative Verification}

\author{Zhenwen Liang$^{1*}$, Ye Liu$^{2}$, Tong Niu$^{2}$, Xiangliang Zhang$^{1}$, Yingbo Zhou$^{2}$, Semih Yavuz$^{2}$\\
$^1$University of Notre Dame \,\,\,\, $^2$Salesforce AI \\
\texttt{\{zliang6,xzhang33\}@nd.edu} \\
\texttt{\{yeliu,tniu,yingbo.zhou,syavuz\}@salesforce.com} \vspace{-0.45cm}
}

\iclrfinalcopy 
\begin{document}

\maketitle

\begin{abstract}
Despite significant advancements in the general capability of large language models (LLMs), they continue to struggle with consistent and accurate reasoning, especially in complex tasks such as mathematical and code reasoning. One key limitation is that LLMs are trained primarily on correct solutions, reducing their ability to detect and learn from errors, which hampers their ability to reliably verify and rank outputs. To address this, we scale up the inference-time computation by generating multiple reasoning paths and employing verifiers to assess and rank the generated outputs by correctness. To facilitate this, we introduce a comprehensive dataset consisting of correct and incorrect solutions for math and code tasks, generated by multiple LLMs. This diverse set of solutions enables verifiers to more effectively distinguish and rank correct answers from erroneous outputs. The training methods for building verifiers were selected based on an extensive comparison of existing approaches. Moreover, to leverage the unique strengths of different reasoning strategies, we propose a novel collaborative method integrating Chain-of-Thought (CoT) and Program-of-Thought (PoT) solutions for verification. CoT provides a clear, step-by-step reasoning process that enhances interpretability, while PoT, being executable, offers a precise and error-sensitive validation mechanism. By taking both of their strengths, our approach significantly improves the accuracy and reliability of reasoning verification. Our verifiers, Math-Rev and Code-Rev, demonstrate substantial performance gains to existing LLMs, achieving state-of-the-art results on benchmarks such as GSM8k and MATH and even outperforming GPT-4o with Qwen-72B-Instruct as the reasoner. 
\blfootnote{This work is done during Zhenwen's internship at Salesforce AI.}
\end{abstract}

\section{Introduction}

Large language models \citep{brown2020language,achiam2023gpt,touvron2023llama,touvron2023llama2,jiang2023mistral,team2024gemma} have demonstrated exceptional performance across various natural language tasks. Notably, the reasoning tasks such as math problem solving 
 \citep{cobbe2021training,hendrycks2021measuring}, code completion \citep{austin2021program,chen2021evaluating}, multi-modal reasoning \citep{yue2024mmmu,liang2024scemqa} have attracted significant attention from AI researchers. Since reasoning is a critical component of many important high-level tasks, such as scientific discovery \citep{liang2024scemqa,miret2024llms}, world model \citep{hao2023reasoning}, embodied agents \citep{song2023llm}, etc. However, even the most advanced LLMs still face challenges in complex multi-step reasoning problems \citep{zhang2024accessing,shi2024can,trinh2024solving}. To improve the performance of LLMs on reasoning, recent studies~\citep{yu2023metamath,yue2023mammoth,gou2023tora,luo2023wizardmath,wei2024magicoder,tang2024mathscale,yue2024mammoth2} have mainly focused on generating synthetic question-answering pairs from stronger LLMs like GPT-4 \citep{achiam2023gpt} or utilizing human-annotated rationales~\citep{toshniwal2024openmathinstruct} for supervised fine-tuning. These approaches have achieved outstanding performance on reasoning benchmarks like GSM8k \citep{cobbe2021training}, MATH \citep{hendrycks2021measuring,lightman2023let}, MBPP \citep{austin2021program}, etc. 

\begin{wrapfigure}[18]{r}{0.48\textwidth}
    \centering
    \includegraphics[width=\linewidth]{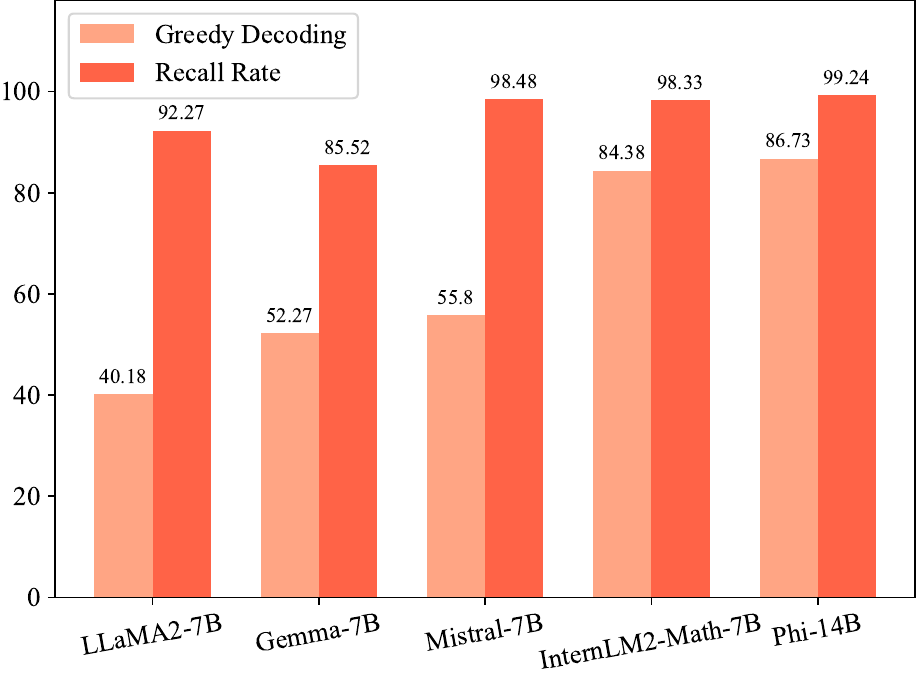}
    \vspace{-0.6cm}
    \caption{Comparison of greedy decoding accuracy and recall out of 64 sampled solutions on GSM8k dataset with various LLMs.}
    \label{fig:gap}     \vspace{-0.5cm}
\end{wrapfigure}
While these straightforward data generation methods have  proven effective,   these LLMs are primarily trained to produce outputs that align with the correct reasoning steps they encountered during training. However, they lack a fundamental understanding of when and why a particular reasoning step might be flawed.
As a result, while LLMs can effectively mimic the structure of correct reasoning paths, they often struggle to ensure the accuracy of these paths and may produce responses that seem correct at first glance, but are flawed \cite{liang2024mathchat}. 
This limitation poses challenges for reliably generating the correct solution. As shown in Fig. \ref{fig:gap}, many LLMs have low accuracy when attempting to find a single solution using greedy decoding. However, when allowing each model to generate 64 solutions (at different temperature settings), the correct answer is often found among the sampled solutions, with a recall rate exceeding 85\%. A similar high recall rate has also been observed by \citep{li2024common}, where models like LLaMA2-7b-base \citep{touvron2023llama2}, despite not being particularly strong in complex reasoning, demonstrate high recall on solving math problems.


This offers hope for addressing the reasoning challenges of LLMs: 
scaling up the inference compute by sampling multiple candidate solutions has emerged as a promising approach  and recently garnered significant attention \citep{zhang2024generative,brown2024large,bansal2024smaller}. Rather than relying solely on the greedy decoding output, these methods involve generating multiple solutions for a given problem by altering the generation temperature or prompt, scoring each solution by a verifier, and selecting the best one with the highest score. Such sample-then-rank strategies can significantly enhance both the accuracy and reliability of   LLM outputs. However, prior studies often focus on specific datasets (e.g., MATH \citep{lightman2023let,wang2023math}) or particular backbone generators (e.g., LLaMA \citep{hosseini2024v} or Gemini \citep{luo2024improve}), leading to the development of ad-hoc verifiers tailored to certain cases. In other words, these verifiers are not generally applicable across different tasks and backbone reasoners,  restricting the generalizability and robustness of LLMs in reasoning and raising difficulties for verifiers in identifying errors \cite{snell2024scaling}.

In this paper, aiming at building verifiers for more effective inference-time verification, we introduce a comprehensive training dataset created by sampling outputs from multiple LLM reasoners of varying  sizes and purposes. We then categorize them into correct and incorrect sets, and use them to build verifiers that learn from the diverse solution patterns produced by different LLMs. Since the methods for training verifiers are so crucial, we conduct a thorough comparison of two key approaches: outcome reward models (ORMs) \citep{cobbe2021training} and preference tuning (e.g., DPO \citep{rafailov2024direct}). ORMs add extra computational heads with scalar outputs to the per-token logits of LLMs and train the model with a binary classification loss. In contrast, preference tuning methods like DPO teach LLMs to learn from pairwise data and generate outputs that align more closely with preferred responses. While preference-tuned LLMs cannot directly output scalar scores like ORMs, we can calculate the likelihood of generating certain solutions given the input problem as the score of the solutions. Our experiments show that reference-free preference tuning methods, such as SimPO \citep{meng2024simpo}, are the most effective for training verifiers. The resulting verifiers for math reasoning and code reasoning are named \textbf{Math} \textbf{R}easoning \textbf{E}nsembled \textbf{V}erifier (\textbf{Math-Rev}) and \textbf{Code} \textbf{R}easoning \textbf{E}nsembled \textbf{V}erifier (\textbf{Code-Rev}) in this paper, respectively.

Moreover, based on our observation, we also explore the complementary strengths of step-by-step language-based solutions and code-based programming solutions for verification purposes. Step-by-step language solutions, also known as chain-of-thought (CoT) \citep{wei2022chain} format, are more descriptive and connected to natural language. In contrast, program solutions, or program-of-thought (PoT) \citep{chen2023program} format, are highly abstract and structured, allowing for direct execution to identify runtime errors, but they are more complex and difficult to read. To address these challenges and leverage the strengths of both formats, we propose a method named \textbf{CoTnPoT} that combines language and code answers during solution verification. Our findings indicate that CoT solutions, being more readable and interpretable by LLMs, enable verifiers to achieve higher performance. On the other hand, code-based solutions, which are executable and sensitive to errors, provide a critical signal when assessing the correctness of language solutions. 

With CoTnPoT and Math-Rev, we achieve significantly better math reasoning verification performance than two baselines - Math-Shepard \citep{wang2023math} and Math-Minos \citep{gao2024llmcriticshelpcatch}. Additionally, as shown in Figure \ref{fig:qwen_math}, our Math-Rev + Qwen2-72B-Instruct outperforms state-of-the-art LLMs including LLaMA3.1-405B and GPT-4o. In summary, our contributions are threefold:
\begin{wrapfigure}{r}{0.4\textwidth}
    \centering
    \includegraphics[width=\linewidth]{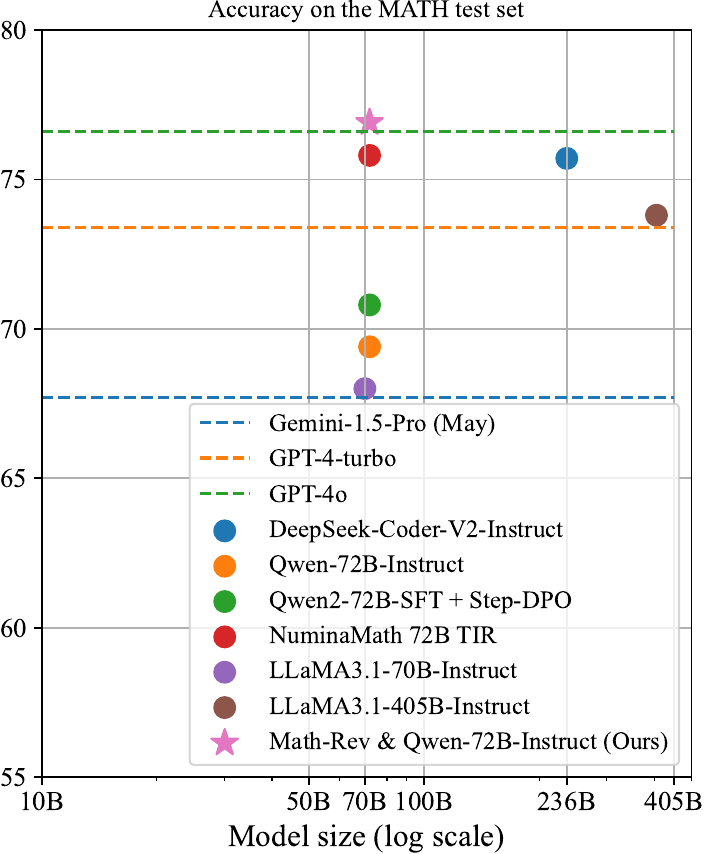}
    \vspace{-0.6cm}
    \caption{Accuracy on the MATH test set across models.}
    \vspace{-0.5cm}
    \label{fig:qwen_math}  
\end{wrapfigure}
\begin{itemize}[leftmargin=0.5em]
\item We develop inference-time verifiers to enhance the reasoning capabilities of LLMs. These verifiers are trained on a comprehensive dataset comprising both correct and incorrect solutions for math and code reasoning tasks. By releasing this dataset, we aim to support the community's efforts in scaling up inference-time computation and advancing research in verifier training and evaluation. 
\item We investigate various verifier training methods and establish that reference-free alignment methods are the most effective. Using SimPO, our developed Math-Rev and Code-Rev achieve state-of-the-art accuracy and surpassing the performance of existing verifiers.
\item We propose a novel method that combines language and code answers for solution verification, achieving promising synchronization and further improving final accuracy. Using Qwen-72B-Instruct \citep{yang2024qwen2} as the backbone reasoner, our approach yields 95.6\% and 76.9\% accuracy on the GSM8k and MATH benchmarks, respectively.
\end{itemize}

\section{Our Method}

\begin{figure}[t]
  \centering
   \includegraphics[width=1.0\textwidth]{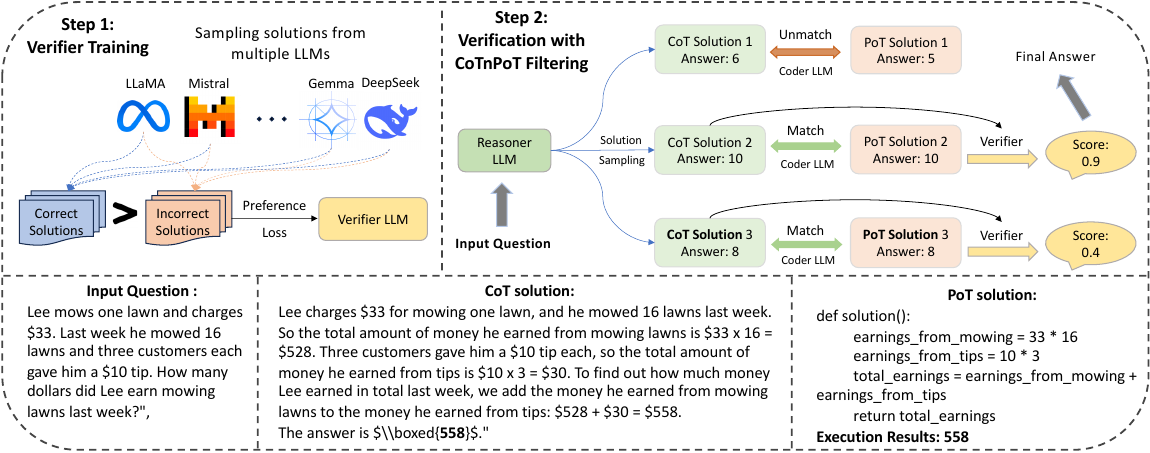}
  \caption{The workflow of our method. We first sample solutions from multiple LLM reasoners and then train verifiers using preference loss (Step 1). During inference (Step 2), we sample multiple CoT solutions per question and use a coder LLM to transform them into a PoT format. Then we filter out any CoT answers that do not match with their corresponding PoT results and feed the remaining CoT solutions to the verifier. The solution with the highest score is selected as the final answer. An example of CoT and PoT solutions is attached.}
  \vspace{-0.4cm}
  \label{fig:method}
\end{figure}

The workflow of our method is presented in Fig. \ref{fig:method}. After collecting a diverse set of solutions, including both correct and incorrect ones, we train our verifiers, which can be implemented using any open-weight LLM (e.g., Mistral-7B-instruct-v0.3). During the inference stage, the reasoner LLM generates responses to an input question, and the verifier is applied to score multiple sampled solutions from the reasoner.


\subsection{Data Collection for Training Verifiers}
\paragraph{Math Reasoning}
We use the training sets of GSM8k \citep{cobbe2021training} and MATH \citep{hendrycks2021measuring} as seed datasets and sample model solutions from multiple backbone models: (1) general-purpose LLMs, including Mistral \citep{jiang2023mistral} and Phi3 \citep{abdin2024phi}; and (2) math-specialized models, including InternLM2-Math \citep{ying2024internlm} and MAmmoTH2-plus \citep{yue2024mammoth2}. For each question in GSM8k and MATH, we sample 10 Chain-of-Thought (CoT) solutions and remove duplicates. Using functions provided by \citep{ying2024internlm}, we extract answers from model predictions and compare them with ground truth, resulting in 159,778 correct and 100,794 incorrect solutions for the training of Math-Rev, with an average of 10.67 correct and 6.73 incorrect solutions per problem. For the evaluation on the MATH dataset, we follow \cite{lightman2023let} and use the subset - MATH500, the same as previous work 
\cite{wang2023math,gao2024llmcriticshelpcatch}. For the results in Figure \ref{fig:qwen_math}, we use the full test set of MATH to ensure fairness. 

\paragraph{Code Reasoning}
Similarly, we utilize general-purpose LLMs, including LLaMA-3-8B \citep{touvron2023llama2} and Phi3 \citep{abdin2024phi}, and code-specialized models, including CodeGemma-7B-it \citep{team2024codegemma} and CodeQwen1.5 \citep{codeqwen1.5}. We select the training sets of MBPP \citep{austin2021program} and the Python subset of MagiCoder-75k \citep{wei2024magicoder} as seed datasets. In code generation tasks, test cases are usually required to determine the correctness of solutions. The original MBPP training set includes test cases, but the MagiCoder does not. To address this, we use GPT-4o to generate test cases for each problem in the Python subset of MagiCoder-75k, retaining only test cases that the reference solution passed. If no generated test case matches the reference solution, we repeat the process with a temperature of 0.8 up to three times. This process results in 11,527 problems with test cases in the MagiCoder-75k dataset. We then generate 50 solutions for each seed problem in both that subset and MBPP, resulting in 132,089 correct and 145,345 incorrect solutions with an average of 11.10 correct and 12.21 incorrect solutions per problem, which are used for training our Code-Rev.

\subsection{Training Math-Rev and Code-Rev}
The verifiers, implemented using LLMs (e.g., Mistral), need to be trained with appropriate training methods to ensure their effectiveness during inference. We extensively investigate various usable methods that are introduced next. 

\paragraph{Reward-based: ORMs and PRMs.}
Following the widely accepted definition in \citep{uesato2022solving}, there are two categories of reward-based methods for building verifiers: outcome-reward models (ORMs) \citep{cobbe2021training} and process-reward models (PRMs) \citep{lightman2023let}. ORM, commonly used in RLHF \citep{ouyang2022training}, can produce scalar scores on model responses, whereas PRM evaluates the reasoning path step-by-step. Despite better performance, PRMs need to collect process supervision data, relying on either human annotation \citep{lightman2023let} or per-step Monte Carlo estimation \citep{wang2023math}, both of which are prohibitively expensive to scale. Moreover, the PRM method requires the solution to be formatted as step-by-step reasoning chains \citep{lightman2023let,wang2023math,luo2024improve}, where steps need to be clearly separated by special tokens or periods to be scored, thereby limiting the application scenario of PRM. Consequently, in this paper, we do not assign per-step scores on reasoning paths, but instead calculate a final score for the whole solution.

\paragraph{Preference-tuning: DPO and Beyond.}
Direct Preference Optimization (DPO) \citep{rafailov2024direct} is one of the most popular offline preference optimization methods. Unlike ORM or PRM which rely on learning an explicit reward model, DPO proposes a novel loss function based on preference pairs, which reparameterizes the reward function and applies it into the the Bradley-Terry (BT) ranking objective. This innovation has inspired various follow-up studies, such as IPO \citep{azar2024general}, KTO \citep{ethayarajh2024kto}, CPO \citep{xu2024contrastive}, and R-DPO \citep{gallego2024refined}. Besides them, the reference-free variants including ORPO \citep{hong2024orpo} and SimPO \citep{meng2024simpo} argue that reference models in the above reward functions would incur additional memory and computational costs and create discrepancy between the reward function and the generation metric during inference.

\paragraph{Our Verifiers Training.}
Although those reference-tuning methods are primarily designated to align LLMs with human preferences, they can also be adapted for training verifiers \citep{hosseini2024v}. By feeding the backbone LLM of the verifiers with pairs of correct and incorrect solutions, designated as chosen and rejected outputs, and applying the mentioned training methods, the verifier can be trained to assign higher generation probabilities to correct solutions over incorrect ones. Then the probability can be served as a score for ranking solutions. In our paper, Math-Rev and Code-Rev are trained separately by their respective training data with one of the preference-tuning methods - SimPO. We believe that such verifiers have a significant advantage over ORMs: it does not introduce additional training parameters and not change the goal of generation for LLMs, aligning better with the original usage of LLM.

\subsection{Inference Enhanced by Verification with CoTnPoT}
\label{sec:cotnpot}
During the inference stage, after deploying our Math-Rev and Code-Rev verifiers, we identify distinct challenges in verifying math and code reasoning. For math reasoning, while model-based verifiers can effectively detect surface-level logical errors such as incorrect use of operators, numbers, and methods, they struggle to catch subtle mistakes such as calculation errors and deeper logical inconsistencies. In code reasoning, the structured and abstract nature of code makes it difficult to read and understand, leading verifiers to assign similar scores to different solutions, indicating their difficulty in accurately identifying errors within the code.

To address these challenges, we propose a method called CoTnPoT, which enhances verification by leveraging the connection and complementary strengths of the Chain of Thought (CoT) and Program of Thought (PoT) solution formats.

For math reasoning, we use an external LLM, DeepseekV2-chat-Lite \citep{zhu2024deepseek}, to transform CoT solutions $S_{CoT}$ into PoT counterparts $S_{PoT}$ based on problem descriptions $Q$, 
\begin{equation}\label{coderval}
S_{PoT} = CoderLLM(Q, S_{CoT}).
\end{equation}
We choose DeepseekV2-chat-Lite because it obtains both strong math reasoning and coding capabilities and we need to apply them to translate CoT solutions into PoT programs for math problems. We then verify whether the transformed final answer from the execution of $S_{PoT}$ matches the final answer from $S_{CoT}$. Our motivation is that logical errors in $S_{CoT}$ would cause run-time errors in $S_{PoT}$, while calculation errors in $S_{CoT}$ would result in mismatched answers between $S_{CoT}$ and $S_{PoT}$, as PoT solutions ensure calculation correctness by using the Python interpreter. This approach takes advantage of the executable nature of program-based solutions.

For code reasoning tasks, we find that directly training verifiers on Python code alone leads to inferior performance. This may be due to the increased difficulty in reading and understanding code compared to human language, which can make it harder to detect reasoning errors. Therefore, we use the same LLM to generate both the code solution $S_{PoT}$ and the corresponding step-by-step description $S_{Des}$ that explains why the solution is correct. Because using the same LLMs for both code and description generation reduces over-reliance on external LLMs. During both training and inference and code verification, we concatenate the description and the code as an integrated input for the verifier, as shown in Equation \ref{coderval_2}. This method provides richer information in the code solutions, making the LLM-based verification process more effective.
\begin{equation}\label{coderval_2}
S_{Des} = CoderLLM(Q, S_{PoT})
\end{equation}

\section{Experiments}

\subsection{Exploring Different Training Methods for Verifiers}

\begin{wrapfigure}{r}{0.4\textwidth}
\vspace{-0.5cm}
    \centering
    \includegraphics[width=0.4\textwidth]{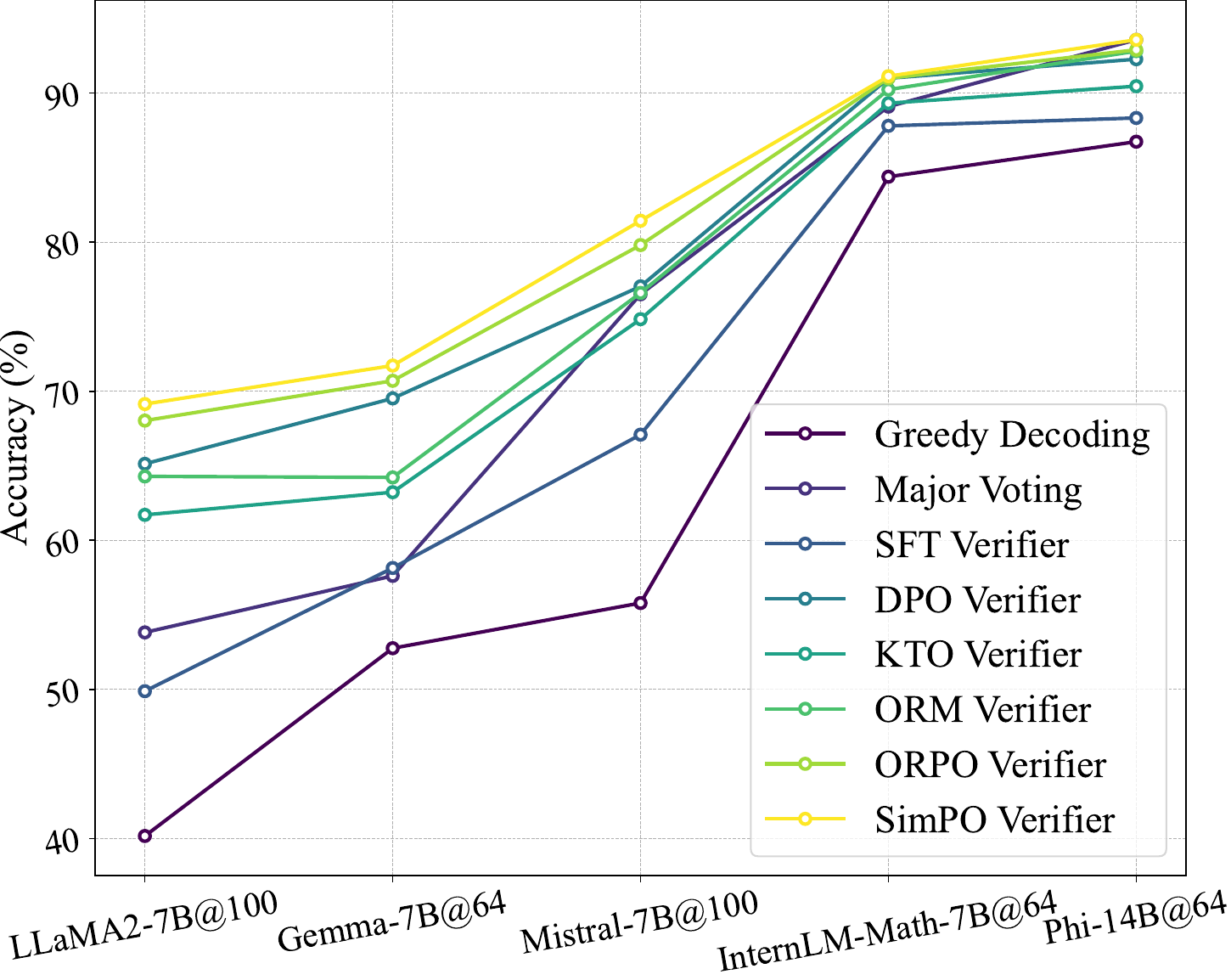}
    \vspace{-0.5cm}
    \caption{Performance of different verifiers (all better than greedy decoding)}
    \label{fig:verifier}
    \vspace{-0.3cm}
\end{wrapfigure}

\paragraph{Experiment Setting.}
For all experiments in Figure \ref{fig:verifier}, we use the latest Mistral-7B-instruct-v0.3 as the backbone LLM for building the verifiers and apply LoRA with a dropout rate of 0.1 to reduce the computational load during verifier training. The training batch size is set to 64, and the learning rate to 0.00002 for all verifiers. For ORM, we add an additional computational head on the per-token logits from the backbone LLM, outputting a scalar value for each token. We take the score of the last token as the final score, which has shown better performance than averaging them based on our observations. For DPO and its variants, we construct preference pairs by randomly selecting correct-incorrect solutions for the same problem from the training set. We use 8 A100-40G GPUs  for all the experiments and employ vLLM to optimize the inference speed. The training of the verifiers takes 5 hours approximately.
We first perform supervised fine-tuning on all correct solutions and then apply preference loss on the preference set.

\paragraph{LLM Reasoners in Evaluation.}
To evaluate the reasoning performance on the GSM8k dataset, we use LLaMA2-7B-base and Mistral-7B-v0.1, both fine-tuned on GSM8k, along with Gemma-7B-it, Phi-14B, InternLM2-Math-7B, and LLaMA3-70B as our reasoners. For LLaMA2 and Mistral, we sample 100 solutions per problem for voting and verification, while 64 solutions are generated for the rest. On the MATH dataset, which contains much harder problems than GSM8k, we replace LLaMA2-7B-base and Mistral-7B-v0.1 with LLaMA3-8B-instruct and Mistral-7B-v0.3  for their superior reasoning ability, along with other four reasoners. For all problems in MATH500, we generate 64 solutions individually. All LLM output sampling in our paper is based on a temperature of 0.8 and top-p of 0.95.

\paragraph{Experimental Results.}

The results are shown in Figure \ref{fig:verifier}. We observe that the verifiers consistently improve the greedy decoding baseline, especially for weaker reasoners such as LLaMA2-7B. 
We also evaluate in-distribution (ID) LLMs, which are the source LLMs used to generate the training data for verifiers, such as Mistral, InternLM2-Math, and Phi, and out-of-distribution (OOD) LLMs, such as LLaMA2-7B and Gemma-7B. The results show no significant difference between ID and OOD performance improvement by verifiers, suggesting that our approach can extend to any LLM reasoners and is not limited to the LLMs that generate the training data. Furthermore, preference-tuning-based verifiers, including DPO and SimPO, outperform ORM, similar to the findings in \cite{hosseini2024v}. The potential reason is that DPO and SimPO train LLMs without changing their structure, thus aligning better with their previous training goals of auto-regressive text generation. Additionally, ORPO and SimPO consistently outperform DPO, potentially because the regularization term on the reference model in the DPO loss might negatively impact verifier training. In other words, we do not need to control the divergence of the SFT model and the final verifier because it will not be used for text generation anymore. Therefore, we can conclude that the reference-free method is more suitable for verifier training.

Additionally, preference-tuning methods such as DPO and SimPO theoretically enable auto-regressive LLMs to generating solutions. However, we observe that the generation ability of verifiers trained with preference pairs degrades rapidly, rendering them incapable of generating coherent sentences. This observation is also consistent with the findings in \cite{hosseini2024v}. We attribute this degradation to that the verifier training process involves more steps and larger learning rates than typical alignment practices, which likely causes the verifier's weights to diverge significantly from the fine-tuned checkpoint. Consequently, these verifiers lose their generation capability and are instead better suited for calculating the likelihood of pre-generated solutions.

\subsection{Evaluation of Verifiers with CoTnPoT}

\input{./tables/coderfilter}

This section focuses on evaluating the inference performance using the trained verifiers with the designed CoTnPoT filtering. We upgraded the backbone model of our verifier in math reasoning from Mistral-7B to MAmmoTH-7B-plus to enhance performance.
\paragraph{Math Reasoning.}
We further enhance the inference process by combining majority voting with verifier scores,  using the scores from verifiers as weights in the voting process. Specifically, we apply Gumbel Softmax \citep{jang2022categorical} with the hyperparameter $\tau$ to regulate the influence of verifier-based scores, as shown in Equation \ref{eq:softmax}.
\begin{equation}\label{eq:softmax}
y_i = \frac{\exp\left(\frac{\log(\pi_i)}{\tau}\right)}{\sum_{j=1}^k \exp\left(\frac{\log(\pi_j)}{\tau}\right)}
\end{equation}
where $\pi_i$ represents the unnormalized log probabilities for the $i$-th solution. Theoretically, if $\tau$ is set to an infinitely large value, the weighted voting will be equivalent to majority voting. If $\tau$ is close to zero, the result will depend solely on the verifier scores. We perform a grid search on $\tau$ values from the set \{0.1, 0.5, 1, 5, 10\} for GSM8k and MATH datasets separately, finding that 0.5 works best for GSM8k and 10 works best for MATH. This implies that for simpler problems like those in GSM8k, we can rely more heavily on verifiers, while for more complex datasets like MATH, the original model outputs should be weighted more significantly.

As shown in Table \ref{tab:performance}, blue percentages indicate performance improvements over the baseline without CoTnPoT, and green percentages indicate improvements over greedy decoding. Generally, we observe that the final column, Weighted Voting + CoTnPoT, consistently outperforms all baselines across all reasoners. CoTnPoT brings improvements to most backbone reasoners and both datasets, demonstrating its effectiveness in filtering incorrect solutions. Notably, CoTnPoT provides a substantial performance boost for weaker reasoners but is less impactful as the reasoners become stronger. This is reasonable because verifying and filtering solutions for strong LLMs is a more challenging task compared to for weaker ones.

\paragraph{Code Reasoning.}
In addition to using PoT to verify and filter CoT answers, we also explore leveraging CoT comments to improve code solution verification.

\input{tables/code.tex}

As shown in Table \ref{tab:code_results}, incorporating CoTnPoT comments into the verification process leads to significant improvements across all LLM reasoners. We believe that the generated comments enrich the information within the solution, enhancing the verifier's understanding of the solution. An ablation study was conducted on the additional training set, i.e., MagiCoder-75k. The experiments show that MagiCoder-75k serves as a valuable additional training resource for coding benchmarks like MBPP. Moreover, we observe that greedy decoding is already a strong baseline for coding tasks, and our verifier-based approaches usually fall short, likely due to the abstractness and obscureness of codes. That is also the reason why our proposed CoTnPoT-based strategy is effective, i.e., we provide high-granularity explanations to clarify the solutions.

\subsection{Comparison with Verifier Baselines and State-of-the-Art Reasoners}

\input{tables/metamath.tex}

We compare our math verifier, Math-Rev, with two recent baselines, Math-Shepard and Math-Minos. We follow their methodology and use a consistent LLM reasoner, MetaMath-7B-Mistral. Although there is a slight difference in that we sampled 64 solutions per problem whereas they sampled 256 solutions, our verifier Math-Rev still achieves the best performance, as shown in Table \ref{tab:mistral_results}. This success is attributed to the more effective verifier training method, SimPO, and the pairwise training data sampled from multiple LLM reasoners. Another notable finding is that our CoTnPoT method poses a slightly negative impact on the MATH500 dataset, the reason is that CoTnPoT is less helpful on stronger backbone reasoners, as also shown in Table \ref{tab:performance}. However, it does not hinder its general applicability demonstrated in Table \ref{tab:performance} and still has the potential to improve by switching the coder model that translates CoT to PoT to stronger ones.

We also pair our Math-Rev with one of the strongest open models, Qwen-72B-Instruct. As shown in Figure \ref{fig:qwen_math}, the final performance of Qwen-72B + Math-Rev on MATH surpasses all SOTA baselines including GPT-4o. This experiment demonstrates that Math-Rev can enhance even the most powerful LLM reasoners, despite being trained on data from smaller and weaker models, highlighting the promising effectiveness of verification - learning from errors.

\subsection{Ablation Study on CoTnPoT}
\begin{wrapfigure}{r}{0.6\textwidth}
  \centering
   \includegraphics[width=0.6\textwidth]{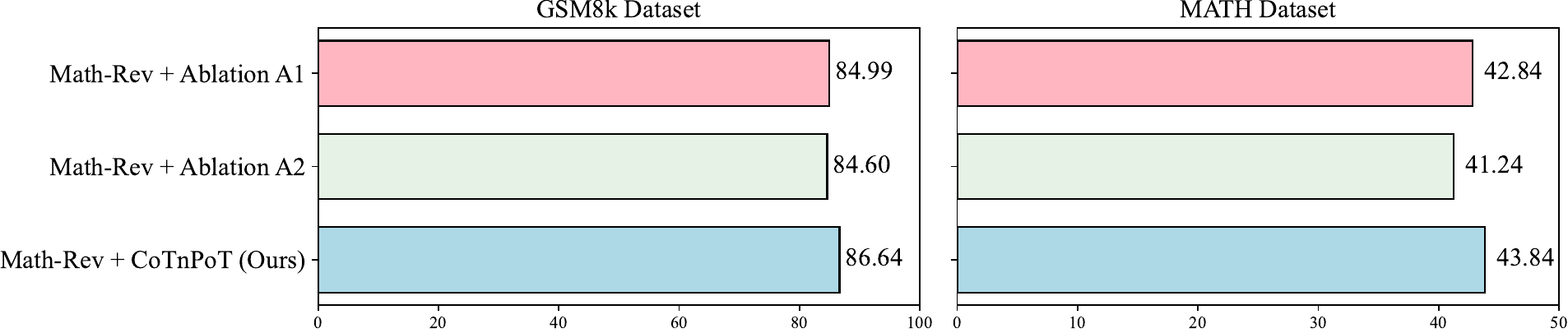}
  \caption{Ablation study on CoTnPoT.}
  \label{fig:ablation}
    \vspace{-0.3cm}
\end{wrapfigure}
In this section, we compare our proposed CoTnPoT with two ablated approaches:

A1. Prompting the same coder LLM to generate the final answer directly through code, and filtering out CoT solutions that do not match the code solution. This ablation isolates the scenario where the coder LLM relies solely on its inherent strong math problem-solving ability, instead of analyzing and transforming the CoT solution.

A2. Prompting the same coder LLM to generate comments that analyze the CoT solutions and assess their correctness. This approach intuitively leverages LLMs as filters for verification.

We implement and compare CoTnPoT, A1, and A2 across all settings and both datasets in Figure \ref{fig:ablation}. The accuracy is averaged at the dataset level for better visibility. We observe that CoTnPoT consistently outperforms both A1 and A2. The potential reason is that the task of translating CoT solutions to PoT solutions is easier and requires less reasoning than the processes in A1 and A2. Therefore, although A1 and A2 are more direct methods to verify a solution, their performance is limited by the capability of the coder LLM. On the other hand, CoTnPoT relies less on complex reasoning, making it more effective overall.


\subsection{Analysis on CoTnPoT}
Our method, CoTnPoT, for math reasoning is designed to filter out low-quality solutions by examining the match between CoT and PoT solutions. This approach essentially functions as a binary classification task. By defining the ground truth label of a correct CoT solution as 1 and an incorrect CoT solution as 0, the correspondence between CoT and PoT solutions is used as the prediction label, where a match is labeled as 1 and a mismatch as 0. The effectiveness of the CoTnPoT filter is directly correlated to the performance of this binary classifier, aiming to retain all solutions labeled as 1 and discard those labeled as 0.

To validate this method, we randomly selected 50,000 correct and 50,000 incorrect CoT solutions from our verifier training set and applied the CoTnPoT filter. The performance of the classifier is summarized in the confusion matrix presented in Table \ref{tab:confusion}. The results demonstrate that the CoTnPoT classifier effectively identifies correct solutions, as evidenced by high True Positive Rate (TPR) and False Negative Rate (FNR). While the False Positive Rate (FPR) and True Negative Rate (TNR) are moderate, indicating some incorrect solutions are not filtered out, the majority of correct solutions are preserved for further verification. This experiment provides strong evidence of the significant performance improvement that the CoTnPoT-based filter brings to math reasoning. Figure \ref{fig:case_math} in the appendix shows the examples of true positive, false negative, false negative, and true negatives of the above CoTnPoT classifier. 
\begin{table}[h!]
\small
    \centering
     \caption{Confusion Matrix for the CoTnPoT-based filter.}
    \vspace{-0.2cm}
    \begin{tabular}{|c|c|c|}
        \hline
        & \makecell{\textbf{Actually Positive:}\\Correct CoT Solution} & \makecell{\textbf{Actually Negative:}\\Wrong CoT Solution} \\
        \hline
         \makecell{\textbf{Predicted Positive:}\\CoTnPoT Match} & True Positives (TPR): 90.09\% & False Positives (FPR): 20.30\% \\
        \hline
        \makecell{\textbf{Predicted Negative:}\\CoTnPoT Mismatch} & False Negatives (FNR): 9.91\% & True Negatives (TNR): 79.70 \\
        \hline
    \end{tabular}
   
    \label{tab:confusion}
\end{table}



\section{Related Work}
\subsection{Scaling up Inference-Time Computing}
\cite{cobbe2021training} is the pioneering work that applies verifiers in mathematical reasoning, where they train token-level reward models to give scores on problem solutions. Then \cite{uesato2022solving,lightman2023let} dive into the application of PRM - process reward models, where scores are assigned to each intermediate step of solutions, providing more fine-grained feedback. Math-Shepherd \citep{wang2023math} and MiPS \citep{wang2024multi} propose using Monte-Carlo Tree-Search (MCTS) to automate the data
collection process instead of human labeling. OVM \cite{yu2024ovm} employs outcome supervision for training a value model, which prioritizes steps that lead to accurate conclusions during inference. V-Star \citep{hosseini2024v} presents an iterative framework in LLM training, which collects both correct data for supervised fine-tuning and wrong data for verifier training. They also showed that DPO is stronger than ORMs in verification. 
Built on reranking strategies such as verifiers, multiple studies \cite{brown2024large,snell2024scaling} found that scaling up inference-time computing is much more cost-effective than training.
To achieve more effective and efficient inference-time verification, our approach samples solutions from various LLM reasoners and comprehensively compares different verifier training methods. Our best verifier Math-Rev achieves strong performance on math solution verification using only outcome-based labels in training and even outperforms PRM baselines.

\subsection{Connect between Chain-of-Thought and Program-of-Thought}
PAL \citep{gao2023pal} and PoT \citep{chen2023program} are two early studies that incorporate Python programs into LLM reasoning. MathCoder \citep{wang2024mathcoder} proposes a method of generating novel and high-quality datasets with math problems and their code-based solutions. As for the code-based verification and feedback, \cite{zhou2024solving} employs a zero-shot prompt on GPT-4 Code Interpreter to encourage it to use code to self-verify its answers. \cite{zhou2024don} autoformalizes informal mathematical statements into formal Isabelle code to verify the internal consistency. ART \citep{miao2024improving} introduces relation tuples into the reasoning steps and verifies them with code interpreter to provide feedback, finally improving reasoning accuracy. Different from these papers, we are the first to investigate the effectiveness of combining CoT and PoT solutions in verification and show promising results on both mathematical and code reasoning tasks.

\section{Conclusion}
In this paper, we address the challenge of improving reasoning verification in LLM by integrating CoT and Program-of-Thought PoT. Firstly, we collect a comprehensive binary dataset, derived from multiple LLM reasoners for both math and code reasoning tasks, providing a robust foundation for training verifiers. Next, through an extensive comparison of outcome reward models (ORMs) and preference-tuning methods, we identify that reference-free preference tuning, particularly SimPO, offers superior performance. Moreover, we introduce techniques to generate CoT/PoT based on their PoT/CoT counterparts for further verification. Our resulting verifiers, Math-Rev and Code-Rev, outperform existing baselines and achieve state-of-the-art results on benchmarks such as GSM8k and MATH. We believe this paper could serve as a strong baseline in reasoning verification and facilitate future studies on reasoning, verifying, reinforcement learning and related areas.

\paragraph{Limitation}
While our approach demonstrates significant improvements in reasoning verification, it also comes with certain limitations. First, the sampling and re-ranking strategy introduces additional computational overhead compared to greedy decoding, which can be resource-intensive, especially when applied to large-scale datasets or deployed in real-time applications. Secondly, our verifier is based on an outcome reward model (ORM) that provides feedback at the solution level rather than at the step level. This solution-level granularity, while effective in overall verification, lacks the finer granularity of process reward models (PRMs) that evaluate each step of the reasoning path. PRMs can potentially offer more detailed feedback and facilitate more precise corrections, particularly in complex multi-step reasoning tasks. However, implementing step-level verification would require extensive process supervision data, which is expensive and challenging to scale.


\bibliography{iclr2024_conference}
\bibliographystyle{iclr2024_conference}

\newpage
\appendix
\section{Appendix}

\subsection{Qualitative Analysis: Error Detection in Solutions}
\begin{figure}[h]
\centering
\includegraphics[width=0.98\textwidth]{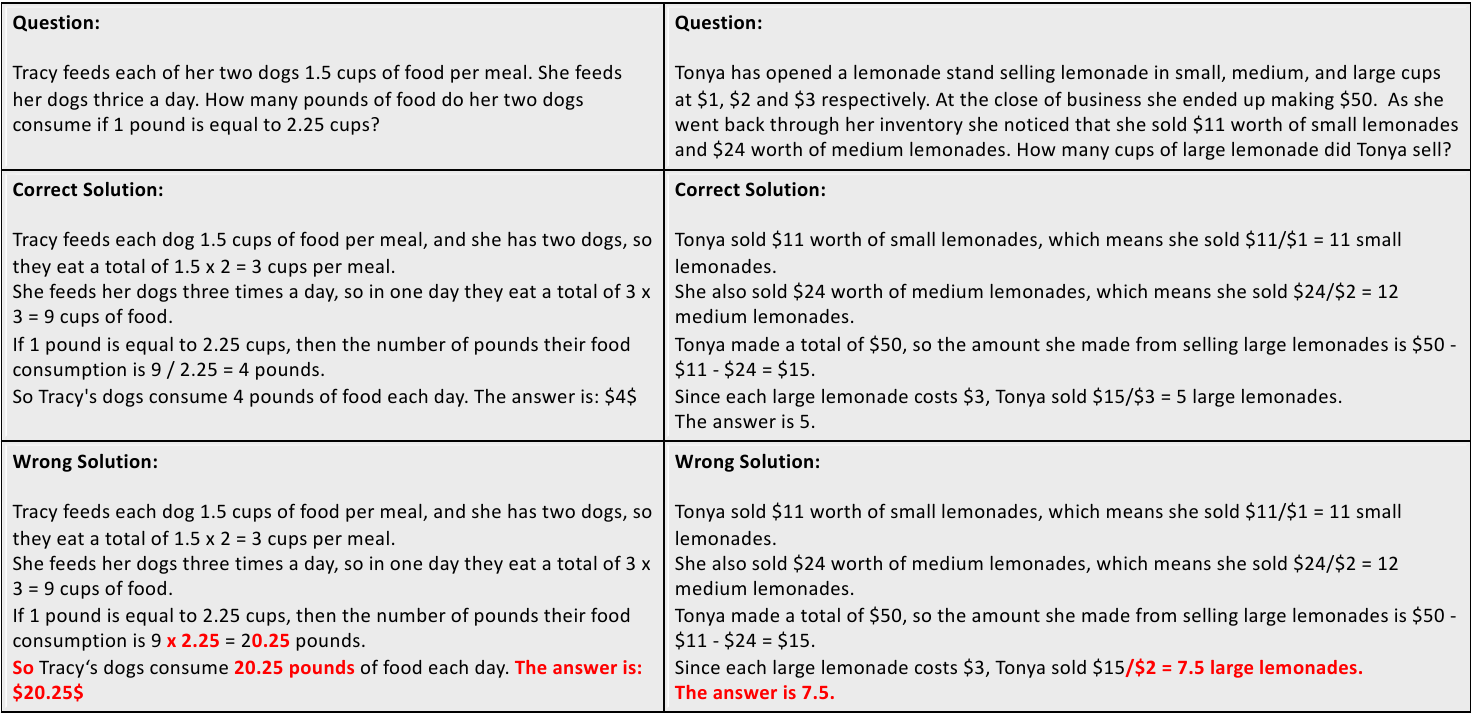}
\caption{The figure illustrates two mathematical problems, each paired with a correct solution and an intentionally incorrect solution. The Math-Rev verifier highlights errors in the incorrect solutions in red, effectively identifying and marking the parts with low log probabilities.}
\vspace{-0.4cm}
\label{fig:heatmap}
\end{figure}

In this experiment, we evaluated the performance of our Math-Rev verifier in identifying and highlighting errors in mathematical solutions. Each column in the provided figure represents a math problem, including both a correct solution and a deliberately modified incorrect solution. We input both solutions into our Math-Rev verifier, and highlight tokens in the wrong solution with log probabilities less than -10 in red to indicate detected errors, as shown in Figure \ref{fig:heatmap}.

For the first problem, Math-Rev successfully identified the incorrect use of the multiplication operator and also recognized the incorrect final answer, highlighting these segments in red. This indicates the verifier's sensitivity to mathematical operations and the final conclusion drawn from these operations. In the second problem, the verifier detected the discrepancy in the calculations and identified the deviation from the problem's requirements, marking the erroneous parts accordingly. This demonstrates Math-Rev's effectiveness in pinpointing computational errors and inconsistencies with problem statements.


\end{document}

%% file: math_commands.tex
\usepackage{amsmath,amsfonts,bm}

\def\eqref#1{equation~\ref{#1}}

\def\1{\bm{1}}

\DeclareMathAlphabet{\mathsfit}{\encodingdefault}{\sfdefault}{m}{sl}
\SetMathAlphabet{\mathsfit}{bold}{\encodingdefault}{\sfdefault}{bx}{n}



%% file: tables/coderfilter.tex
\begin{table}[t]
\small
    \centering
    \caption{Performance improvement brought by the proposed CoTnPoT. The best performance each row is highlighted. \textcolor{OliveGreen}{Green arrow} denotes the percentage improvement over greedy decoding, \textcolor{positive}{blue arrow} indicates the improvement over the baseline without CoTnPoT.}
    \vspace{-0.3cm}
    \resizebox{0.99\textwidth}{!}{
    \begin{tabularx}{1.05\textwidth}{lcccccc}
\toprule
 & \makecell{Sampling +\\CoTnPoT} & \makecell{Voting + \\CoTnPoT} & \makecell{Recall +\\CoTnPoT} & SimPO & \makecell{SimPO +\\CoTnPoT} & \makecell{Weighted \\Voting + \\CoTnPoT} \\
\midrule
\multicolumn{7}{l}{\textbf{GSM8k:}} \\[0.2em]
\hspace{0.3em}LLaMA2-7B-GSM8k & 56.56 & 67.25 & 88.48 & 75.21 & 78.01 & \textbf{78.09} \\
& \textcolor{positive}{↑ 40.77\%} & \textcolor{positive}{↑ 24.93\%} & \textcolor{negative}{↓ 4.11\%} & \textcolor{black}{0\%} & \textcolor{positive}{↑ 3.72\%} & \textcolor{positive}{↑ 3.66\%} \\
& \textcolor{OliveGreen}{↑ 40.77\%} & \textcolor{OliveGreen}{↑ 67.37\%} & \textcolor{OliveGreen}{↑ 120.21\%} & \textcolor{OliveGreen}{↑ 87.18\%} & \textcolor{OliveGreen}{↑ 94.15\%} & \textcolor{OliveGreen}{↑ 94.35\%} \\

\hspace{0.3em}Mistral-7B-GSM8k & 71.34 & 84.76 & 96.66 & 87.87 & 89.54 & \textbf{89.69} \\
& \textcolor{positive}{↑ 27.85\%} & \textcolor{positive}{↑ 10.80\%} & \textcolor{negative}{↓ 1.85\%} & \textcolor{black}{0\%} & \textcolor{positive}{↑ 1.90\%} & \textcolor{positive}{↑ 1.94\%} \\
& \textcolor{OliveGreen}{↑ 27.85\%} & \textcolor{OliveGreen}{↑ 51.90\%} & \textcolor{OliveGreen}{↑ 73.23\%} & \textcolor{OliveGreen}{↑ 57.47\%} & \textcolor{OliveGreen}{↑ 60.47\%} & \textcolor{OliveGreen}{↑ 60.73\%} \\

\hspace{0.3em}Gemma-7B-it & 66.79 & 71.11 & 83.62 & 75.06 & \textbf{78.54} & \textbf{78.54} \\
& \textcolor{positive}{↑ 26.57\%} & \textcolor{positive}{↑ 23.41\%} & \textcolor{negative}{↓ 2.22\%} & \textcolor{black}{0\%} & \textcolor{positive}{↑ 4.64\%} & \textcolor{positive}{↑ 4.58\%} \\
& \textcolor{OliveGreen}{↑ 26.57\%} & \textcolor{OliveGreen}{↑ 34.75\%} & \textcolor{OliveGreen}{↑ 58.46\%} & \textcolor{OliveGreen}{↑ 42.24\%} & \textcolor{OliveGreen}{↑ 48.83\%} & \textcolor{OliveGreen}{↑ 48.83\%} \\

\hspace{0.3em}InternLM2-Math-7B & 88.40 & 91.21 & 97.42 & 92.34 & 92.49 & \textbf{92.65} \\
& \textcolor{positive}{↑ 4.76\%} & \textcolor{positive}{↑ 2.39\%} & \textcolor{negative}{↓ 0.93\%} & \textcolor{black}{0\%} & \textcolor{positive}{↑ 0.16\%} & \textcolor{positive}{↑ 0.23\%} \\
& \textcolor{OliveGreen}{↑ 4.76\%} & \textcolor{OliveGreen}{↑ 8.09\%} & \textcolor{OliveGreen}{↑ 15.45\%} & \textcolor{OliveGreen}{↑ 9.43\%} & \textcolor{OliveGreen}{↑ 9.61\%} & \textcolor{OliveGreen}{↑ 9.80\%} \\

\hspace{0.3em}Phi3-14B & 89.99 & 94.19 & 99.01 & 94.16 & 94.47 & \textbf{94.62} \\
& \textcolor{positive}{↑ 3.76\%} & \textcolor{positive}{↑ 0.67\%} & \textcolor{negative}{↓ 0.23\%} & \textcolor{black}{0\%} & \textcolor{positive}{↑ 0.33\%} & \textcolor{positive}{↑ 0.45\%} \\
& \textcolor{OliveGreen}{↑ 3.76\%} & \textcolor{OliveGreen}{↑ 8.60\%} & \textcolor{OliveGreen}{↑ 14.16\%} & \textcolor{OliveGreen}{↑ 8.57\%} & \textcolor{OliveGreen}{↑ 8.92\%} & \textcolor{OliveGreen}{↑ 9.10\%} \\

\hspace{0.3em}LLaMA3-70B-instruct & 94.92 & 95.45 & 97.73 & 95.22 & 95.30 & \textbf{95.60} \\
& \textcolor{positive}{↑ 0.56\%} & \textcolor{positive}{↑ 0.24\%} & \textcolor{negative}{↓ 0.76\%} & \textcolor{black}{0\%} & \textcolor{positive}{↑ 0.08\%} & \textcolor{positive}{↑ 0.33\%} \\
& \textcolor{OliveGreen}{↑ 0.56\%} & \textcolor{OliveGreen}{↑ 1.12\%} & \textcolor{OliveGreen}{↑ 3.54\%} & \textcolor{OliveGreen}{↑ 0.88\%} & \textcolor{OliveGreen}{↑ 0.96\%} & \textcolor{OliveGreen}{↑ 1.28\%} \\

\midrule
\multicolumn{7}{l}{\textbf{MATH500:}} \\[0.2em]
\hspace{0.3em}LLaMA3-8B-Instruct & 40.20 & 41.60 & 63.60 & 45.00 & 45.80 & \textbf{46.00} \\
& \textcolor{positive}{↑ 34.00\%} & \textcolor{positive}{↑ 13.04\%} & \textcolor{negative}{↓ 8.88\%} & \textcolor{black}{0\%} & \textcolor{positive}{↑ 1.78\%} & \textcolor{positive}{↑ 1.77\%} \\
& \textcolor{OliveGreen}{↑ 34.00\%} & \textcolor{OliveGreen}{↑ 38.67\%} & \textcolor{OliveGreen}{↑ 112.00\%} & \textcolor{OliveGreen}{↑ 50.00\%} & \textcolor{OliveGreen}{↑ 52.67\%} & \textcolor{OliveGreen}{↑ 53.33\%} \\

\hspace{0.3em}Mistral-Instruct-v0.3 & 28.40 & 32.40 & 50.00 & 32.60 & 35.40 & \textbf{35.60} \\
& \textcolor{positive}{↑ 121.87\%} & \textcolor{positive}{↑ 54.29\%} & \textcolor{negative}{↓ 13.79\%} & \textcolor{black}{0\%} & \textcolor{positive}{↑ 8.59\%} & \textcolor{positive}{↑ 7.88\%} \\
& \textcolor{OliveGreen}{↑ 121.87\%} & \textcolor{OliveGreen}{↑ 153.12\%} & \textcolor{OliveGreen}{↑ 290.62\%} & \textcolor{OliveGreen}{↑ 154.69\%} & \textcolor{OliveGreen}{↑ 176.56\%} & \textcolor{OliveGreen}{↑ 178.12\%} \\

\hspace{0.3em}Gemma-7B-it & 33.20 & 35.80 & 51.60 & 32.80 & 39.20 & \textbf{39.60} \\
& \textcolor{positive}{↑ 104.94\%} & \textcolor{positive}{↑ 50.42\%} & \textcolor{negative}{↓ 9.79\%} & \textcolor{black}{0\%} & \textcolor{positive}{↑ 19.51\%} & \textcolor{positive}{↑ 18.56\%} \\
& \textcolor{OliveGreen}{↑ 104.94\%} & \textcolor{OliveGreen}{↑ 120.99\%} & \textcolor{OliveGreen}{↑ 218.52\%} & \textcolor{OliveGreen}{↑ 102.47\%} & \textcolor{OliveGreen}{↑ 141.98\%} & \textcolor{OliveGreen}{↑ 144.44\%} \\

\hspace{0.3em}InternLM2-Math-7B & 58.20 & 63.00 & 76.00 & 62.00 & 63.60 & \textbf{63.80} \\
& \textcolor{positive}{↑ 62.57\%} & \textcolor{positive}{↑ 12.90\%} & \textcolor{negative}{↓ 2.31\%} & \textcolor{black}{0\%} & \textcolor{positive}{↑ 2.58\%} & \textcolor{positive}{↑ 2.24\%} \\
& \textcolor{OliveGreen}{↑ 62.57\%} & \textcolor{OliveGreen}{↑ 75.98\%} & \textcolor{OliveGreen}{↑ 112.29\%} & \textcolor{OliveGreen}{↑ 73.18\%} & \textcolor{OliveGreen}{↑ 77.65\%} & \textcolor{OliveGreen}{↑ 78.21\%} \\

\hspace{0.3em}Phi3-14B & 42.80 & 48.20 & 65.00 & 50.80 & 50.00 & \textbf{50.20} \\
& \textcolor{positive}{↑ 81.36\%} & \textcolor{positive}{↑ 4.78\%} & \textcolor{negative}{↓ 11.92\%} & \textcolor{black}{0\%} & \textcolor{negative}{↓ 1.57\%} & \textcolor{negative}{↓ 1.18\%} \\
& \textcolor{OliveGreen}{↑ 81.36\%} & \textcolor{OliveGreen}{↑ 104.24\%} & \textcolor{OliveGreen}{↑ 175.42\%} & \textcolor{OliveGreen}{↑ 115.25\%} & \textcolor{OliveGreen}{↑ 111.86\%} & \textcolor{OliveGreen}{↑ 112.71\%} \\

\hspace{0.3em}LLaMA3-70B-instruct & 56.80 & 61.20 & 76.00 & 56.80 & 60.80 & \textbf{62.80} \\
& \textcolor{positive}{↑ 9.23\%} & \textcolor{positive}{↑ 3.38\%} & \textcolor{negative}{↓ 12.64\%} & \textcolor{black}{0\%} & \textcolor{positive}{↑ 7.04\%} & \textcolor{positive}{↑ 8.28\%} \\
& \textcolor{OliveGreen}{↑ 9.23\%} & \textcolor{OliveGreen}{↑ 17.69\%} & \textcolor{OliveGreen}{↑ 46.15\%} & \textcolor{OliveGreen}{↑ 9.23\%} & \textcolor{OliveGreen}{↑ 16.92\%} & \textcolor{OliveGreen}{↑ 20.77\%} \\

\bottomrule
    \end{tabularx}}
    \vspace{-0.5cm}
    \label{tab:performance}
\end{table}

%% file: tables/code.tex
\begin{table}
\footnotesize
\centering
\renewcommand{\arraystretch}{1.1}
\caption{Performance of different verification strategies on Code-Rev. We compare the performance on using the MBPP training set alone and incorporating MagiCoder, and the verification on code solution only and solution with CoTnPoT comments. Left and right numbers are top-1 pass rates on MBPP and MBPP+, respectively. The \textcolor{OliveGreen}{green arrows} denote the percentage change compared to greedy decoding performance.}
\vspace{-0.3cm}
\resizebox{1\textwidth}{!}{
\begin{tabular}{cccccc}
\toprule
 & Codegemma & Phi & LLaMA3 & CodeQwen & DeepseekCoder \\ 
\midrule
\multirow{2}{*}{\makecell{MBPP \\ w/o CoTnPoT}} & 64.2/53.9 & 72.2/58.3 & 60.4/51.2 & 75.7/65.7 & 72.0/60.8 \\ 
& \textcolor{negative}{↓ 8.81\%} / \textcolor{negative}{↓ 5.27\%} & \textcolor{OliveGreen}{↑ 0.14\%} / \textcolor{OliveGreen}{↑ 1.04\%} & \textcolor{negative}{↓ 13.84\%} / \textcolor{negative}{↓ 13.66\%} & \textcolor{negative}{↓ 4.66\%} / \textcolor{negative}{↓ 4.78\%} & \textcolor{negative}{↓ 4.26\%} / \textcolor{negative}{↓ 2.25\%} \\ 
\midrule
\multirow{2}{*}{\makecell{MBPP \\ w CoTnPoT}} & 67.6/55.4 & 74.9/60.0 & 66.2/54.8 & 79.5/69.6 & 73.9/62.6 \\ 
& \textcolor{negative}{↓ 3.98\%} / \textcolor{negative}{↓ 2.64\%} & \textcolor{OliveGreen}{↑ 3.88\%} / \textcolor{OliveGreen}{↑ 3.99\%} & \textcolor{negative}{↓ 5.56\%} / \textcolor{negative}{↓ 7.59\%} & \textcolor{OliveGreen}{↑ 0.13\%} / \textcolor{OliveGreen}{↑ 0.87\%} & \textcolor{negative}{↓ 1.73\%} / \textcolor{OliveGreen}{↑ 0.64\%} \\ 
\midrule
\multirow{2}{*}{\makecell{MBPP + MagiCoder \\ w/o CoTnPoT}} & 65.1/54.8 & 73.7/58.4 & 63.3/52.6 & 77.5/66.5 & 73.0/62.2 \\ 
& \textcolor{negative}{↓ 7.53\%} / \textcolor{negative}{↓ 3.69\%} & \textcolor{OliveGreen}{↑ 2.22\%} / \textcolor{OliveGreen}{↑ 1.21\%} & \textcolor{negative}{↓ 9.70\%} / \textcolor{negative}{↓ 11.30\%} & \textcolor{negative}{↓ 2.39\%} / \textcolor{negative}{↓ 3.62\%} & \textcolor{negative}{↓ 2.93\%} / \textcolor{OliveGreen}{0.00\%} \\ 
\midrule
\multirow{2}{*}{\makecell{MBPP + MagiCoder \\ w CoTnPoT}} & \textbf{70.9/58.3} & \textbf{75.2/60.5} & \textbf{72.7/62.0} & \textbf{80.3/71.1} & \textbf{77.5/67.3} \\ 
& \textcolor{OliveGreen}{↑ 0.71\%} / \textcolor{OliveGreen}{↑ 2.46\%} & \textcolor{OliveGreen}{↑ 4.30\%} / \textcolor{OliveGreen}{↑ 4.85\%} & \textcolor{OliveGreen}{↑ 3.71\%} / \textcolor{OliveGreen}{↑ 4.55\%} & \textcolor{OliveGreen}{↑ 1.13\%} / \textcolor{OliveGreen}{↑ 3.04\%} & \textcolor{OliveGreen}{↑ 3.06\%} / \textcolor{OliveGreen}{↑ 8.20\%} \\ 

\bottomrule
\end{tabular}
}

\vspace{-0.4cm}
\label{tab:code_results}
\end{table}

%% file: tables/metamath.tex
\begin{table}[h]
\footnotesize
    \centering
    \caption{Our verifier Math-Rev outperforms two baselines with fewer solutions sampled per problem on both GSM8k and Math500 datasets, demonstrating the effectiveness of our verifier training and CoTnPoT verification.}
    \vspace{-0.3cm}
    \begin{tabularx}{0.80\textwidth}{lcc}
        \toprule
        \textit{\textbf{Mistral-7B-MetaMath Results}} & GSM8k & MATH500 \\
        \midrule
        Major Voting @ 64 & 83.50 & 35.00 \\
        Major Voting @ 256 & 83.90 & 35.10 \\
        Math-Shepherd @ 256 \citep{wang2023math} & 87.10 & 37.30 \\
        Math-Shepherd + Voting @ 256 \citep{wang2023math} & 86.30 & 38.30 \\
        ORM + PPO + Voting @ 256 \citep{wang2023math} & 89.00 & 43.10 \\
        Math-Shepherd + PPO + Voting @ 256 \citep{wang2023math} & 89.10 & 43.50 \\
        Math-Minos (ORM) @ 256 \citep{gao2024llmcriticshelpcatch} & 87.30 & 37.40 \\
        Math-Minos (PRM) @ 256 \citep{gao2024llmcriticshelpcatch} & 87.60 & 37.80 \\
        Math-Minos (ORM) + Voting @ 256 \citep{gao2024llmcriticshelpcatch}  & 88.20 & 38.30 \\
        Math-Minos (PRM) + Voting @ 256 \citep{gao2024llmcriticshelpcatch} & 87.80 & 38.60 \\
        Math-Rev (Ours) @ 64 & 90.37 & \textbf{46.60} \\
        Math-Rev + CoTnPoT (Ours) @ 64 & \textbf{90.75} & 46.40 \\
        \bottomrule
    \end{tabularx}
    \vspace{-0.4cm}
    \label{tab:mistral_results}
\end{table}

%% file: iclr2024_conference.bbl
\begin{thebibliography}{62}
\providecommand{\natexlab}[1]{#1}
\providecommand{\url}[1]{\texttt{#1}}
\expandafter\ifx\csname urlstyle\endcsname\relax
  \providecommand{\doi}[1]{doi: #1}\else
  \providecommand{\doi}{doi: \begingroup \urlstyle{rm}\Url}\fi

\bibitem[Abdin et~al.(2024)Abdin, Jacobs, Awan, Aneja, Awadallah, Awadalla, Bach, Bahree, Bakhtiari, Behl, et~al.]{abdin2024phi}
Marah Abdin, Sam~Ade Jacobs, Ammar~Ahmad Awan, Jyoti Aneja, Ahmed Awadallah, Hany Awadalla, Nguyen Bach, Amit Bahree, Arash Bakhtiari, Harkirat Behl, et~al.
\newblock Phi-3 technical report: A highly capable language model locally on your phone.
\newblock \emph{arXiv preprint arXiv:2404.14219}, 2024.

\bibitem[Achiam et~al.(2023)Achiam, Adler, Agarwal, Ahmad, Akkaya, Aleman, Almeida, Altenschmidt, Altman, Anadkat, et~al.]{achiam2023gpt}
Josh Achiam, Steven Adler, Sandhini Agarwal, Lama Ahmad, Ilge Akkaya, Florencia~Leoni Aleman, Diogo Almeida, Janko Altenschmidt, Sam Altman, Shyamal Anadkat, et~al.
\newblock Gpt-4 technical report.
\newblock \emph{arXiv preprint arXiv:2303.08774}, 2023.

\bibitem[Austin et~al.(2021)Austin, Odena, Nye, Bosma, Michalewski, Dohan, Jiang, Cai, Terry, Le, et~al.]{austin2021program}
Jacob Austin, Augustus Odena, Maxwell Nye, Maarten Bosma, Henryk Michalewski, David Dohan, Ellen Jiang, Carrie Cai, Michael Terry, Quoc Le, et~al.
\newblock Program synthesis with large language models.
\newblock \emph{arXiv preprint arXiv:2108.07732}, 2021.

\bibitem[Azar et~al.(2024)Azar, Guo, Piot, Munos, Rowland, Valko, and Calandriello]{azar2024general}
Mohammad~Gheshlaghi Azar, Zhaohan~Daniel Guo, Bilal Piot, Remi Munos, Mark Rowland, Michal Valko, and Daniele Calandriello.
\newblock A general theoretical paradigm to understand learning from human preferences.
\newblock In \emph{International Conference on Artificial Intelligence and Statistics}, pp.\  4447--4455. PMLR, 2024.

\bibitem[Bansal et~al.(2024)Bansal, Hosseini, Agarwal, Tran, and Kazemi]{bansal2024smaller}
Hritik Bansal, Arian Hosseini, Rishabh Agarwal, Vinh~Q Tran, and Mehran Kazemi.
\newblock Smaller, weaker, yet better: Training llm reasoners via compute-optimal sampling.
\newblock \emph{arXiv preprint arXiv:2408.16737}, 2024.

\bibitem[Brown et~al.(2024)Brown, Juravsky, Ehrlich, Clark, Le, R{\'e}, and Mirhoseini]{brown2024large}
Bradley Brown, Jordan Juravsky, Ryan Ehrlich, Ronald Clark, Quoc~V Le, Christopher R{\'e}, and Azalia Mirhoseini.
\newblock Large language monkeys: Scaling inference compute with repeated sampling.
\newblock \emph{arXiv preprint arXiv:2407.21787}, 2024.

\bibitem[Brown et~al.(2020)Brown, Mann, Ryder, Subbiah, Kaplan, Dhariwal, Neelakantan, Shyam, Sastry, Askell, et~al.]{brown2020language}
Tom Brown, Benjamin Mann, Nick Ryder, Melanie Subbiah, Jared~D Kaplan, Prafulla Dhariwal, Arvind Neelakantan, Pranav Shyam, Girish Sastry, Amanda Askell, et~al.
\newblock Language models are few-shot learners.
\newblock \emph{Advances in neural information processing systems}, 33:\penalty0 1877--1901, 2020.

\bibitem[Chen et~al.(2021)Chen, Tworek, Jun, Yuan, Pinto, Kaplan, Edwards, Burda, Joseph, Brockman, et~al.]{chen2021evaluating}
Mark Chen, Jerry Tworek, Heewoo Jun, Qiming Yuan, Henrique Ponde De~Oliveira Pinto, Jared Kaplan, Harri Edwards, Yuri Burda, Nicholas Joseph, Greg Brockman, et~al.
\newblock Evaluating large language models trained on code.
\newblock \emph{arXiv preprint arXiv:2107.03374}, 2021.

\bibitem[Chen et~al.(2023)Chen, Ma, Wang, and Cohen]{chen2023program}
Wenhu Chen, Xueguang Ma, Xinyi Wang, and William~W Cohen.
\newblock Program of thoughts prompting: Disentangling computation from reasoning for numerical reasoning tasks.
\newblock \emph{Transactions on Machine Learning Research}, 2023.

\bibitem[Cobbe et~al.(2021)Cobbe, Kosaraju, Bavarian, Chen, Jun, Kaiser, Plappert, Tworek, Hilton, Nakano, et~al.]{cobbe2021training}
Karl Cobbe, Vineet Kosaraju, Mohammad Bavarian, Mark Chen, Heewoo Jun, Lukasz Kaiser, Matthias Plappert, Jerry Tworek, Jacob Hilton, Reiichiro Nakano, et~al.
\newblock Training verifiers to solve math word problems.
\newblock \emph{arXiv preprint arXiv:2110.14168}, 2021.

\bibitem[Ethayarajh et~al.(2024)Ethayarajh, Xu, Muennighoff, Jurafsky, and Kiela]{ethayarajh2024kto}
Kawin Ethayarajh, Winnie Xu, Niklas Muennighoff, Dan Jurafsky, and Douwe Kiela.
\newblock Kto: Model alignment as prospect theoretic optimization.
\newblock \emph{arXiv preprint arXiv:2402.01306}, 2024.

\bibitem[Gallego(2024)]{gallego2024refined}
V{\'\i}ctor Gallego.
\newblock Refined direct preference optimization with synthetic data for behavioral alignment of llms.
\newblock \emph{arXiv preprint arXiv:2402.08005}, 2024.

\bibitem[Gao et~al.(2024)Gao, Cai, Xu, Wang, Zheng, Lin, Lu, Liu, Zhou, Xiao, Hu, Liu, and Chang]{gao2024llmcriticshelpcatch}
Bofei Gao, Zefan Cai, Runxin Xu, Peiyi Wang, Ce~Zheng, Runji Lin, Keming Lu, Dayiheng Liu, Chang Zhou, Wen Xiao, Junjie Hu, Tianyu Liu, and Baobao Chang.
\newblock Llm critics help catch bugs in mathematics: Towards a better mathematical verifier with natural language feedback, 2024.
\newblock URL \url{https://arxiv.org/abs/2406.14024}.

\bibitem[Gao et~al.(2023)Gao, Madaan, Zhou, Alon, Liu, Yang, Callan, and Neubig]{gao2023pal}
Luyu Gao, Aman Madaan, Shuyan Zhou, Uri Alon, Pengfei Liu, Yiming Yang, Jamie Callan, and Graham Neubig.
\newblock Pal: Program-aided language models.
\newblock In \emph{International Conference on Machine Learning}, pp.\  10764--10799. PMLR, 2023.

\bibitem[Gou et~al.(2024)Gou, Shao, Gong, Yang, Huang, Duan, Chen, et~al.]{gou2023tora}
Zhibin Gou, Zhihong Shao, Yeyun Gong, Yujiu Yang, Minlie Huang, Nan Duan, Weizhu Chen, et~al.
\newblock Tora: A tool-integrated reasoning agent for mathematical problem solving.
\newblock \emph{The Twelfth International Conference on Learning Representations}, 2024.

\bibitem[Hao et~al.(2023)Hao, Gu, Ma, Hong, Wang, Wang, and Hu]{hao2023reasoning}
Shibo Hao, Yi~Gu, Haodi Ma, Joshua~Jiahua Hong, Zhen Wang, Daisy~Zhe Wang, and Zhiting Hu.
\newblock Reasoning with language model is planning with world model.
\newblock In \emph{The 2023 Conference on Empirical Methods in Natural Language Processing}, 2023.

\bibitem[Hendrycks et~al.(2021)Hendrycks, Burns, Kadavath, Arora, Basart, Tang, Song, and Steinhardt]{hendrycks2021measuring}
Dan Hendrycks, Collin Burns, Saurav Kadavath, Akul Arora, Steven Basart, Eric Tang, Dawn Song, and Jacob Steinhardt.
\newblock Measuring mathematical problem solving with the math dataset.
\newblock In \emph{Thirty-fifth Conference on Neural Information Processing Systems Datasets and Benchmarks Track (Round 2)}, 2021.

\bibitem[Hong et~al.(2024)Hong, Lee, and Thorne]{hong2024orpo}
Jiwoo Hong, Noah Lee, and James Thorne.
\newblock Orpo: Monolithic preference optimization without reference model.
\newblock \emph{arXiv preprint arXiv:2403.07691}, 2024.

\bibitem[Hosseini et~al.(2024)Hosseini, Yuan, Malkin, Courville, Sordoni, and Agarwal]{hosseini2024v}
Arian Hosseini, Xingdi Yuan, Nikolay Malkin, Aaron Courville, Alessandro Sordoni, and Rishabh Agarwal.
\newblock V-star: Training verifiers for self-taught reasoners.
\newblock \emph{arXiv preprint arXiv:2402.06457}, 2024.

\bibitem[Jang et~al.(2022)Jang, Gu, and Poole]{jang2022categorical}
Eric Jang, Shixiang Gu, and Ben Poole.
\newblock Categorical reparameterization with gumbel-softmax.
\newblock In \emph{International Conference on Learning Representations}, 2022.

\bibitem[Jiang et~al.(2023)Jiang, Sablayrolles, Mensch, Bamford, Chaplot, Casas, Bressand, Lengyel, Lample, Saulnier, et~al.]{jiang2023mistral}
Albert~Q Jiang, Alexandre Sablayrolles, Arthur Mensch, Chris Bamford, Devendra~Singh Chaplot, Diego de~las Casas, Florian Bressand, Gianna Lengyel, Guillaume Lample, Lucile Saulnier, et~al.
\newblock Mistral 7b.
\newblock \emph{arXiv preprint arXiv:2310.06825}, 2023.

\bibitem[Li et~al.(2024)Li, Wang, Hu, Wei, Zheng, Hu, Zhang, and Peng]{li2024common}
Chen Li, Weiqi Wang, Jingcheng Hu, Yixuan Wei, Nanning Zheng, Han Hu, Zheng Zhang, and Houwen Peng.
\newblock Common 7b language models already possess strong math capabilities.
\newblock \emph{arXiv preprint arXiv:2403.04706}, 2024.

\bibitem[Liang et~al.(2024{\natexlab{a}})Liang, Guo, Liu, Guo, Zhou, Yang, Jiao, Pi, Zhang, and Zhang]{liang2024scemqa}
Zhenwen Liang, Kehan Guo, Gang Liu, Taicheng Guo, Yujun Zhou, Tianyu Yang, Jiajun Jiao, Renjie Pi, Jipeng Zhang, and Xiangliang Zhang.
\newblock Scemqa: A scientific college entrance level multimodal question answering benchmark.
\newblock \emph{ACL}, 2024{\natexlab{a}}.

\bibitem[Liang et~al.(2024{\natexlab{b}})Liang, Yu, Yu, Yao, Zhang, Zhang, and Yu]{liang2024mathchat}
Zhenwen Liang, Dian Yu, Wenhao Yu, Wenlin Yao, Zhihan Zhang, Xiangliang Zhang, and Dong Yu.
\newblock Mathchat: Benchmarking mathematical reasoning and instruction following in multi-turn interactions.
\newblock \emph{arXiv preprint arXiv:2405.19444}, 2024{\natexlab{b}}.

\bibitem[Lightman et~al.(2023)Lightman, Kosaraju, Burda, Edwards, Baker, Lee, Leike, Schulman, Sutskever, and Cobbe]{lightman2023let}
Hunter Lightman, Vineet Kosaraju, Yura Burda, Harri Edwards, Bowen Baker, Teddy Lee, Jan Leike, John Schulman, Ilya Sutskever, and Karl Cobbe.
\newblock Let's verify step by step.
\newblock \emph{arXiv preprint arXiv:2305.20050}, 2023.

\bibitem[Luo et~al.(2023)Luo, Sun, Xu, Zhao, Lou, Tao, Geng, Lin, Chen, and Zhang]{luo2023wizardmath}
Haipeng Luo, Qingfeng Sun, Can Xu, Pu~Zhao, Jianguang Lou, Chongyang Tao, Xiubo Geng, Qingwei Lin, Shifeng Chen, and Dongmei Zhang.
\newblock Wizardmath: Empowering mathematical reasoning for large language models via reinforced evol-instruct.
\newblock \emph{arXiv preprint arXiv:2308.09583}, 2023.

\bibitem[Luo et~al.(2024)Luo, Liu, Liu, Phatale, Lara, Li, Shu, Zhu, Meng, Sun, et~al.]{luo2024improve}
Liangchen Luo, Yinxiao Liu, Rosanne Liu, Samrat Phatale, Harsh Lara, Yunxuan Li, Lei Shu, Yun Zhu, Lei Meng, Jiao Sun, et~al.
\newblock Improve mathematical reasoning in language models by automated process supervision.
\newblock \emph{arXiv preprint arXiv:2406.06592}, 2024.

\bibitem[Meng et~al.(2024)Meng, Xia, and Chen]{meng2024simpo}
Yu~Meng, Mengzhou Xia, and Danqi Chen.
\newblock Simpo: Simple preference optimization with a reference-free reward.
\newblock \emph{arXiv preprint arXiv:2405.14734}, 2024.

\bibitem[Miao et~al.(2024)Miao, Zhao, and Tsuruoka]{miao2024improving}
Zhongtao Miao, Kaiyan Zhao, and Yoshimasa Tsuruoka.
\newblock Improving arithmetic reasoning ability of large language models through relation tuples, verification and dynamic feedback.
\newblock \emph{arXiv preprint arXiv:2406.17873}, 2024.

\bibitem[Miret \& Krishnan(2024)Miret and Krishnan]{miret2024llms}
Santiago Miret and NM~Krishnan.
\newblock Are llms ready for real-world materials discovery?
\newblock \emph{arXiv preprint arXiv:2402.05200}, 2024.

\bibitem[Ouyang et~al.(2022)Ouyang, Wu, Jiang, Almeida, Wainwright, Mishkin, Zhang, Agarwal, Slama, Ray, et~al.]{ouyang2022training}
Long Ouyang, Jeffrey Wu, Xu~Jiang, Diogo Almeida, Carroll Wainwright, Pamela Mishkin, Chong Zhang, Sandhini Agarwal, Katarina Slama, Alex Ray, et~al.
\newblock Training language models to follow instructions with human feedback.
\newblock \emph{Advances in neural information processing systems}, 35:\penalty0 27730--27744, 2022.

\bibitem[Rafailov et~al.(2024)Rafailov, Sharma, Mitchell, Manning, Ermon, and Finn]{rafailov2024direct}
Rafael Rafailov, Archit Sharma, Eric Mitchell, Christopher~D Manning, Stefano Ermon, and Chelsea Finn.
\newblock Direct preference optimization: Your language model is secretly a reward model.
\newblock \emph{Advances in Neural Information Processing Systems}, 36, 2024.

\bibitem[Shi et~al.(2024)Shi, Tang, Narasimhan, and Yao]{shi2024can}
Quan Shi, Michael Tang, Karthik Narasimhan, and Shunyu Yao.
\newblock Can language models solve olympiad programming?
\newblock \emph{arXiv preprint arXiv:2404.10952}, 2024.

\bibitem[Snell et~al.(2024)Snell, Lee, Xu, and Kumar]{snell2024scaling}
Charlie Snell, Jaehoon Lee, Kelvin Xu, and Aviral Kumar.
\newblock Scaling llm test-time compute optimally can be more effective than scaling model parameters.
\newblock \emph{arXiv preprint arXiv:2408.03314}, 2024.

\bibitem[Song et~al.(2023)Song, Wu, Washington, Sadler, Chao, and Su]{song2023llm}
Chan~Hee Song, Jiaman Wu, Clayton Washington, Brian~M Sadler, Wei-Lun Chao, and Yu~Su.
\newblock Llm-planner: Few-shot grounded planning for embodied agents with large language models.
\newblock In \emph{Proceedings of the IEEE/CVF International Conference on Computer Vision}, pp.\  2998--3009, 2023.

\bibitem[Tang et~al.(2024)Tang, Zhang, Wan, and Wei]{tang2024mathscale}
Zhengyang Tang, Xingxing Zhang, Benyou Wan, and Furu Wei.
\newblock Mathscale: Scaling instruction tuning for mathematical reasoning.
\newblock \emph{arXiv preprint arXiv:2403.02884}, 2024.

\bibitem[Team(2024{\natexlab{a}})]{team2024codegemma}
CodeGemma Team.
\newblock Codegemma: Open code models based on gemma.
\newblock \emph{arXiv preprint arXiv:2406.11409}, 2024{\natexlab{a}}.

\bibitem[Team et~al.(2024)Team, Mesnard, Hardin, Dadashi, Bhupatiraju, Pathak, Sifre, Rivi{\`e}re, Kale, Love, et~al.]{team2024gemma}
Gemma Team, Thomas Mesnard, Cassidy Hardin, Robert Dadashi, Surya Bhupatiraju, Shreya Pathak, Laurent Sifre, Morgane Rivi{\`e}re, Mihir~Sanjay Kale, Juliette Love, et~al.
\newblock Gemma: Open models based on gemini research and technology.
\newblock \emph{arXiv preprint arXiv:2403.08295}, 2024.

\bibitem[Team(2024{\natexlab{b}})]{codeqwen1.5}
Qwen Team.
\newblock Code with codeqwen1.5, April 2024{\natexlab{b}}.
\newblock URL \url{https://qwenlm.github.io/blog/codeqwen1.5/}.

\bibitem[Toshniwal et~al.(2024)Toshniwal, Moshkov, Narenthiran, Gitman, Jia, and Gitman]{toshniwal2024openmathinstruct}
Shubham Toshniwal, Ivan Moshkov, Sean Narenthiran, Daria Gitman, Fei Jia, and Igor Gitman.
\newblock Openmathinstruct-1: A 1.8 million math instruction tuning dataset.
\newblock \emph{arXiv preprint arXiv:2402.10176}, 2024.

\bibitem[Touvron et~al.(2023{\natexlab{a}})Touvron, Lavril, Izacard, Martinet, Lachaux, Lacroix, Rozi{\`e}re, Goyal, Hambro, Azhar, et~al.]{touvron2023llama}
Hugo Touvron, Thibaut Lavril, Gautier Izacard, Xavier Martinet, Marie-Anne Lachaux, Timoth{\'e}e Lacroix, Baptiste Rozi{\`e}re, Naman Goyal, Eric Hambro, Faisal Azhar, et~al.
\newblock Llama: Open and efficient foundation language models.
\newblock \emph{arXiv preprint arXiv:2302.13971}, 2023{\natexlab{a}}.

\bibitem[Touvron et~al.(2023{\natexlab{b}})Touvron, Martin, Stone, Albert, Almahairi, Babaei, Bashlykov, Batra, Bhargava, Bhosale, et~al.]{touvron2023llama2}
Hugo Touvron, Louis Martin, Kevin Stone, Peter Albert, Amjad Almahairi, Yasmine Babaei, Nikolay Bashlykov, Soumya Batra, Prajjwal Bhargava, Shruti Bhosale, et~al.
\newblock Llama 2: Open foundation and fine-tuned chat models.
\newblock \emph{arXiv preprint arXiv:2307.09288}, 2023{\natexlab{b}}.

\bibitem[Trinh et~al.(2024)Trinh, Wu, Le, He, and Luong]{trinh2024solving}
Trieu~H Trinh, Yuhuai Wu, Quoc~V Le, He~He, and Thang Luong.
\newblock Solving olympiad geometry without human demonstrations.
\newblock \emph{Nature}, 625\penalty0 (7995):\penalty0 476--482, 2024.

\bibitem[Uesato et~al.(2022)Uesato, Kushman, Kumar, Song, Siegel, Wang, Creswell, Irving, and Higgins]{uesato2022solving}
Jonathan Uesato, Nate Kushman, Ramana Kumar, Francis Song, Noah Siegel, Lisa Wang, Antonia Creswell, Geoffrey Irving, and Irina Higgins.
\newblock Solving math word problems with process-and outcome-based feedback.
\newblock \emph{arXiv preprint arXiv:2211.14275}, 2022.

\bibitem[Wang et~al.(2024{\natexlab{a}})Wang, Ren, Zhou, Lu, Luo, Shi, Zhang, Song, Zhan, and Li]{wang2024mathcoder}
Ke~Wang, Houxing Ren, Aojun Zhou, Zimu Lu, Sichun Luo, Weikang Shi, Renrui Zhang, Linqi Song, Mingjie Zhan, and Hongsheng Li.
\newblock Mathcoder: Seamless code integration in llms for enhanced mathematical reasoning.
\newblock In \emph{12th International Conference on Learning Representations (ICLR 2024)}, 2024{\natexlab{a}}.

\bibitem[Wang et~al.(2023)Wang, Li, Shao, Xu, Dai, Li, Chen, Wu, and Sui]{wang2023math}
Peiyi Wang, Lei Li, Zhihong Shao, RX~Xu, Damai Dai, Yifei Li, Deli Chen, Y~Wu, and Zhifang Sui.
\newblock Math-shepherd: A label-free step-by-step verifier for llms in mathematical reasoning.
\newblock \emph{arXiv preprint arXiv:2312.08935}, 2023.

\bibitem[Wang et~al.(2024{\natexlab{b}})Wang, Li, Wu, Luo, Hou, Yu, and Shang]{wang2024multi}
Zihan Wang, Yunxuan Li, Yuexin Wu, Liangchen Luo, Le~Hou, Hongkun Yu, and Jingbo Shang.
\newblock Multi-step problem solving through a verifier: An empirical analysis on model-induced process supervision.
\newblock \emph{arXiv preprint arXiv:2402.02658}, 2024{\natexlab{b}}.

\bibitem[Wei et~al.(2022)Wei, Wang, Schuurmans, Bosma, Xia, Chi, Le, Zhou, et~al.]{wei2022chain}
Jason Wei, Xuezhi Wang, Dale Schuurmans, Maarten Bosma, Fei Xia, Ed~Chi, Quoc~V Le, Denny Zhou, et~al.
\newblock Chain-of-thought prompting elicits reasoning in large language models.
\newblock \emph{NeurIPS}, 35:\penalty0 24824--24837, 2022.

\bibitem[Wei et~al.(2024)Wei, Wang, Liu, Ding, and Zhang]{wei2024magicoder}
Yuxiang Wei, Zhe Wang, Jiawei Liu, Yifeng Ding, and Lingming Zhang.
\newblock Magicoder: Empowering code generation with oss-instruct.
\newblock In \emph{Forty-first International Conference on Machine Learning}, 2024.

\bibitem[Xu et~al.(2024)Xu, Sharaf, Chen, Tan, Shen, Van~Durme, Murray, and Kim]{xu2024contrastive}
Haoran Xu, Amr Sharaf, Yunmo Chen, Weiting Tan, Lingfeng Shen, Benjamin Van~Durme, Kenton Murray, and Young~Jin Kim.
\newblock Contrastive preference optimization: Pushing the boundaries of llm performance in machine translation.
\newblock \emph{arXiv preprint arXiv:2401.08417}, 2024.

\bibitem[Yang et~al.(2024)Yang, Yang, Hui, Zheng, Yu, Zhou, Li, Li, Liu, Huang, et~al.]{yang2024qwen2}
An~Yang, Baosong Yang, Binyuan Hui, Bo~Zheng, Bowen Yu, Chang Zhou, Chengpeng Li, Chengyuan Li, Dayiheng Liu, Fei Huang, et~al.
\newblock Qwen2 technical report.
\newblock \emph{arXiv preprint arXiv:2407.10671}, 2024.

\bibitem[Ying et~al.(2024)Ying, Zhang, Li, Zhou, Shao, Fei, Ma, Hong, Liu, Wang, et~al.]{ying2024internlm}
Huaiyuan Ying, Shuo Zhang, Linyang Li, Zhejian Zhou, Yunfan Shao, Zhaoye Fei, Yichuan Ma, Jiawei Hong, Kuikun Liu, Ziyi Wang, et~al.
\newblock Internlm-math: Open math large language models toward verifiable reasoning.
\newblock \emph{arXiv preprint arXiv:2402.06332}, 2024.

\bibitem[Yu et~al.(2024{\natexlab{a}})Yu, Gao, and Wang]{yu2024ovm}
Fei Yu, Anningzhe Gao, and Benyou Wang.
\newblock Ovm, outcome-supervised value models for planning in mathematical reasoning.
\newblock In \emph{Findings of the Association for Computational Linguistics: NAACL 2024}, pp.\  858--875, 2024{\natexlab{a}}.

\bibitem[Yu et~al.(2024{\natexlab{b}})Yu, Jiang, Shi, Jincheng, Liu, Zhang, Kwok, Li, Weller, and Liu]{yu2023metamath}
Longhui Yu, Weisen Jiang, Han Shi, YU~Jincheng, Zhengying Liu, Yu~Zhang, James Kwok, Zhenguo Li, Adrian Weller, and Weiyang Liu.
\newblock Metamath: Bootstrap your own mathematical questions for large language models.
\newblock In \emph{The Twelfth International Conference on Learning Representations}, 2024{\natexlab{b}}.

\bibitem[Yue et~al.(2024{\natexlab{a}})Yue, Ni, Zhang, Zheng, Liu, Zhang, Stevens, Jiang, Ren, Sun, et~al.]{yue2024mmmu}
Xiang Yue, Yuansheng Ni, Kai Zhang, Tianyu Zheng, Ruoqi Liu, Ge~Zhang, Samuel Stevens, Dongfu Jiang, Weiming Ren, Yuxuan Sun, et~al.
\newblock Mmmu: A massive multi-discipline multimodal understanding and reasoning benchmark for expert agi.
\newblock In \emph{Proceedings of the IEEE/CVF Conference on Computer Vision and Pattern Recognition}, pp.\  9556--9567, 2024{\natexlab{a}}.

\bibitem[Yue et~al.(2024{\natexlab{b}})Yue, Qu, Zhang, Fu, Huang, Sun, Su, and Chen]{yue2023mammoth}
Xiang Yue, Xingwei Qu, Ge~Zhang, Yao Fu, Wenhao Huang, Huan Sun, Yu~Su, and Wenhu Chen.
\newblock Mammoth: Building math generalist models through hybrid instruction tuning.
\newblock In \emph{The Twelfth International Conference on Learning Representations}, 2024{\natexlab{b}}.

\bibitem[Yue et~al.(2024{\natexlab{c}})Yue, Zheng, Zhang, and Chen]{yue2024mammoth2}
Xiang Yue, Tuney Zheng, Ge~Zhang, and Wenhu Chen.
\newblock Mammoth2: Scaling instructions from the web.
\newblock \emph{arXiv preprint arXiv:2405.03548}, 2024{\natexlab{c}}.

\bibitem[Zhang et~al.(2024{\natexlab{a}})Zhang, Li, Huang, Zhou, Li, and Ouyang]{zhang2024accessing}
Di~Zhang, Jiatong Li, Xiaoshui Huang, Dongzhan Zhou, Yuqiang Li, and Wanli Ouyang.
\newblock Accessing gpt-4 level mathematical olympiad solutions via monte carlo tree self-refine with llama-3 8b.
\newblock \emph{arXiv preprint arXiv:2406.07394}, 2024{\natexlab{a}}.

\bibitem[Zhang et~al.(2024{\natexlab{b}})Zhang, Hosseini, Bansal, Kazemi, Kumar, and Agarwal]{zhang2024generative}
Lunjun Zhang, Arian Hosseini, Hritik Bansal, Mehran Kazemi, Aviral Kumar, and Rishabh Agarwal.
\newblock Generative verifiers: Reward modeling as next-token prediction.
\newblock \emph{arXiv preprint arXiv:2408.15240}, 2024{\natexlab{b}}.

\bibitem[Zhou et~al.(2024{\natexlab{a}})Zhou, Wang, Lu, Shi, Luo, Qin, Lu, Jia, Song, Zhan, et~al.]{zhou2024solving}
Aojun Zhou, Ke~Wang, Zimu Lu, Weikang Shi, Sichun Luo, Zipeng Qin, Shaoqing Lu, Anya Jia, Linqi Song, Mingjie Zhan, et~al.
\newblock Solving challenging math word problems using gpt-4 code interpreter with code-based self-verification.
\newblock In \emph{12th International Conference on Learning Representations (ICLR 2024)}, 2024{\natexlab{a}}.

\bibitem[Zhou et~al.(2024{\natexlab{b}})Zhou, Staats, Li, Szegedy, Weinberger, and Wu]{zhou2024don}
Jin~Peng Zhou, Charles Staats, Wenda Li, Christian Szegedy, Kilian~Q Weinberger, and Yuhuai Wu.
\newblock Don't trust: Verify--grounding llm quantitative reasoning with autoformalization.
\newblock \emph{arXiv preprint arXiv:2403.18120}, 2024{\natexlab{b}}.

\bibitem[Zhu et~al.(2024)Zhu, Guo, Shao, Yang, Wang, Xu, Wu, Li, Gao, Ma, et~al.]{zhu2024deepseek}
Qihao Zhu, Daya Guo, Zhihong Shao, Dejian Yang, Peiyi Wang, Runxin Xu, Y~Wu, Yukun Li, Huazuo Gao, Shirong Ma, et~al.
\newblock Deepseek-coder-v2: Breaking the barrier of closed-source models in code intelligence.
\newblock \emph{arXiv preprint arXiv:2406.11931}, 2024.

\end{thebibliography}
